\documentclass[journal]{IEEEtran}
\pdfoutput=1
\usepackage{amssymb}
\usepackage{amsmath}
\usepackage{graphicx}
\usepackage{multirow}
\usepackage{url}
\usepackage[font=small]{caption}
\usepackage{hhline}
\graphicspath{{images/} }

\ifCLASSINFOpdf
\else
\fi
\begin{document}
%
\title{Using a Deep Reinforcement Learning Agent for Traffic Signal Control}
%
%
%

\author{Wade Genders\textsuperscript{a}, Saiedeh Razavi\textsuperscript{b}{}
        {}
        {}
\thanks{\textsuperscript{a} Ph.D. Student, Department of Civil Engineering, McMaster University, Hamilton, Ontario, L8S 4L8 Canada email: genderwt@mcmaster.ca}
\thanks{\textsuperscript{b} Associate Professor, Chair in Heavy Construction, Department of Civil Engineering, McMaster University, Hamilton, Ontario, L8S 4L8 Canada email: razavi@mcmaster.ca}
\thanks{This work has been submitted to the IEEE for possible publication. Copyright may be transferred without notice, after which this version may no longer be accessible.}}

%
%

\markboth{}%
{Shell \MakeLowercase{\textit{et al.}}: Bare Demo of IEEEtran.cls for IEEE Journals}
%



\maketitle

\begin{abstract}
Ensuring transportation systems are efficient is a priority for modern society. Technological advances have made it possible for transportation systems to collect large volumes of varied data on an unprecedented scale. We propose a traffic signal control system which takes advantage of this new, high quality data, with minimal abstraction compared to other proposed systems. We apply modern deep reinforcement learning methods to build a truly adaptive traffic signal control agent in the traffic microsimulator SUMO. We propose a new state space, the discrete traffic state encoding, which is information dense. The discrete traffic state encoding is used as input to a deep convolutional neural network, trained using Q-learning with experience replay. Our agent was compared against a one hidden layer neural network traffic signal control agent and reduces average cumulative delay by 82\%, average queue length by 66\% and average travel time by 20\%.
\end{abstract}


%
\IEEEpeerreviewmaketitle

\section{Introduction} \label{intro}
%
%
%
%
\IEEEPARstart{M}{odern} society relies on its many transportation systems for the movement of individuals, goods and services. Ensuring vehicles can move efficiently from their origin to destination is desirable by all. However, increasing population, and subsequent vehicle ownership, has increased the demand of road infrastructure often beyond its capacity, resulting in congestion, travel delays and unnecessary vehicle emissions. To address this problem, two types of solutions are possible. The first is to increase capacity by expanding road infrastructure, however this can be expensive, protracted and decrease capacity in the short term. The second solution is to increase the efficiency of existing infrastructure and the systems that govern them, such as traffic signal controllers (TSC). We advocate this second solution, by utilizing recent advancements from the domain of artificial intelligence \cite{mnih2015human} to develop a new traffic signal controller. \par We define the traffic signal control problem as follows; given the state of traffic at an intersection, what is the optimal traffic signal phase and sequence that should be enacted? Many systems have been proposed that utilize new sensors, particularly reinforcement learning for traffic signal control, however they do not take full advantage of the available data.  We propose a deep artificial neural network as a traffic signal control agent (TSCA), trained using reinforcement learning, that strives to solve the traffic signal control problem by developing an optimal control policy.\par Reinforcement learning is a machine learning paradigm where an agent seeks to maximize cumulative reward by developing a state-action policy through repeated interaction with its environment. Reinforcement learning agents achieve optimal control with respect to a defined reward by developing an optimal state-action policy. Function approximators, such as artificial neural networks, have been used in reinforcement learning to approximate value functions when the agent's representation of the environment, or state space, becomes too large \cite{tesauro1995temporal}. Convolutional neural networks, a specific type of network architecture, are inspired by biological research on the animal visual cortex \cite{hubel1962receptive}\cite{hubel1968receptive} and have displayed impressive performance \cite{lecun1998gradient}. They apply the mathematical convolution operation between various filters and the layer input to produce feature maps. Convolutional networks are advantageous because minimal input pre-processing is required and they can develop their own features. We develop a deep Q-network traffic signal control agent (DQTSCA), with the action-value function modeled as a deep convolutional neural network trained using reinforcement learning in a traffic microsimulator, SUMO, on an isolated intersection.\par Reinforcement learning is a suitable technique for attempting to solve the traffic signal control problem, as it elegantly represents the elements of the problem - agent (traffic signal controller), environment (state of traffic) and actions (traffic signals). Previous research using reinforcement learning for traffic signal control has yielded impressive results \cite{balaji2010urban}\cite{prashanth2011reinforcement}\cite{el2013multiagent}, yet we perceive areas for improvement. We propose a new state space definition, the discrete traffic state encoding (DTSE), as an improved representation of traffic, as it contains more relevant information compared to previous research's state space definitions. The DTSE is proposed as it is information dense; the convolutional neural network is required to take advantage of the information dense state. The DTSE will allow the convolutional neural network to perceive more relevant traffic information than previous research, extract useful features and develop high-level state representations. The agent can then achieve optimal control by choosing the actions with the highest value, or maximum expected cumulative reward.\par The succeeding sections are organized as follows: Section \ref{lit} details research conducted in the domain of traffic signal control and reinforcement learning, Section \ref{design} describes the proposed DQTSCA and defines the state space, action space and reward, Section \ref{experiment} details the tools used to implement the proposed agent and describes its training, Section \ref{results} discusses the results and performance of the agent and Section \ref{conclusion} summarizes the research conducted and provides ideas for future work.
 
\section{Literature Review} \label{lit}
Significant research has been conducted using reinforcement learning for traffic signal control. Early efforts were limited by simple simulations and a lack of computational power \cite{thorpe1996tra}\cite{wiering2000multi}\cite{brockfeld2001optimizing}\cite{abdulhai2003reinforcement}. Beginning in the early 2000's, continuous improvements in both of these areas have created a variety of simulation tools that are increasingly complex and realistic. Traffic microsimulators are the most popular tool used by traffic researchers, as they model individual vehicles as distinct entities and can reproduce real-world traffic behavior such as shockwaves. Research conducted has differed in reinforcement learning type, state space definition, action space definition, reward definition, simulator, traffic network geometry and vehicle generation model. Previous research efforts have defined the state space as some attribute of traffic, the number of queued vehicles \cite{wiering2000multi}\cite{abdulhai2003reinforcement}\cite{chin2011q}\cite{abdoos2013holonic} and traffic flow \cite{arel2010reinforcement}\cite{balaji2010urban} the most popular. The action space has been defined as all available signal phases \cite{arel2010reinforcement}\cite{el2013multiagent} or restricted to green phases only \cite{balaji2010urban}\cite{chin2011q}\cite{abdoos2013holonic}. The most common reward definitions are change in delay \cite{arel2010reinforcement}\cite{el2013multiagent} and change in queued vehicles \cite{balaji2010urban}\cite{chin2011q}\cite{abdoos2013holonic}. For a comprehensive review of reinforcement learning traffic signal control research, the reader is referred to \cite{el2014design} and \cite{mannion2016experimental}.\par Regarding previous research, the following observations can be made. First, the majority of state definitions are abstractions of the traffic state which omit relevant information. A reinforcement learning agent must first observe the state of the environment before it can act, if useful information is missing, it is unlikely to be able to act optimally. For example, if the state space is defined as the number of queued vehicles at the intersection, this ignores all of the moving vehicles, as well as the queued vehicles' lane and queue position. We believe the state space definition should include as much relevant information about the traffic state as possible, including vehicles' location and speed, thus our proposal of the DTSE, formally defined in Section \ref{design}. We recognize that in practice it may be difficult for a TSCA to observe the state of all vehicles' location and speed, but we will defend this assumption in succeeding sections. However, some previous research has proposed a similar, less abstracted, yet limited, state definition \cite{thorpe1996tra}, from which our research acknowledges their contribution and seeks to extend beyond their efforts.\par Second, the TSCA should be given as much action autonomy as possible, therefore it must be recognized that defining the action space as choosing between fixed sequences of signal phases is limiting. For example, if we define that an advance left green signal phase must always precede a through green signal phase, this assumes the optimal policy follows such a sequence. However, it is conceivable that the optimal action given a certain traffic state is to have an advance left green signal phase succeed a through green signal phase. Much of the previous research has constrained the agent's action in such a way; our action space definition seeks to endow the agent with a higher degree of autonomy in an attempt to learn the optimal policy.\par Finally, all previous research have used computer simulations, as real-world experimentation is infeasible for various reasons.The majority of research assumes vehicle generation can be modeled as a Poisson process, which relies upon the negative exponential distribution, to model the time between vehicle generation events. We propose in subsequent sections that the negative exponential is not the best distribution to model real traffic, as empirical research has shown other distributions to more accurately model different vehicle generation flow rates.

\section{Proposed System} \label{design}
Attempting to solve the traffic signal control problem using reinforcement learning requires a formulation of the problem in the language of reinforcement learning, specifically, defining a state space $S$, an action space $A$ and a reward $R$.
\subsection{State Space}
We propose the DTSE as the appropriate state space $S$ in this research, inspired by a common technique in computing of discretization and quantization of continuous entities. For each lane approaching the intersection, the DTSE discretizes a length $l$ of the lane segment, beginning at the stop line, into cells of length $c$. The selection of $c$ will change the behavior of system. If $c$ is many times larger than the average vehicle length, the individual dynamics of each vehicle will be lost, however computational cost will be reduced. If $c$ is much smaller than the average vehicle length, the individual vehicle dynamics will be retained, however the computational cost will increase, perhaps unnecessarily. We mention the selection of $c$ is important, however for this research we select $c$ in a simplified manner in an attempt to evaluate the proposed system.\par
\begin{figure}
\centering
\includegraphics[scale=0.25]{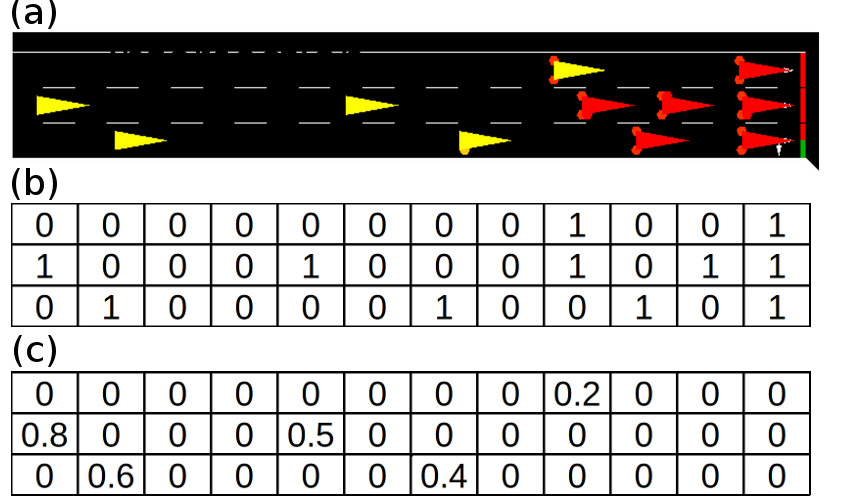}
\caption{Example of simulated traffic (a) with corresponding Boolean- (b) and real-valued DTSE vectors (c).}
\label{fig:dtse}
\end{figure}

 The DTSE is composed of three vectors, the first representing the presence of a vehicle or not in the cell, the second the speed of the vehicle and the third the current traffic signal phase (i.e., the most recent action selected). The addition of second speed vector is an extension beyond \cite{thorpe1996tra}, as their state definition only consists of a vector representing the presence of a vehicle. Therefore, the state of traffic at an intersection with $n$ lanes is formally defined as the DTSE, where $S \in \left({{\mathbb{B}}\times{\mathbb{R}}}\right)^{{{\frac{l}{c}}\times{n}}}\times{P}$ and $P$ represents the current traffic signal phase. At time $t$, the agent observes the traffic state (i.e., the DTSE) as $s_t \in S$. A representation of the DTSE can be seen in Fig. \ref{fig:dtse}, with triangles representing vehicles traveling from left to right. In Fig. \ref{fig:dtse}, Fig. \ref{fig:dtse} (a) shows simulated vehicles approaching the intersection, Fig. \ref{fig:dtse} (b) is the Boolean-valued vector of the DTSE, encoding the presence or absence of a value and Fig. \ref{fig:dtse} (c) is the real-valued vector of the DTSE, encoding the normalized speed.
\par The motivation behind the DTSE is to retain useful information. If the agent is to discover the optimal policy, it must discover the optimal actions for any given state; having knowledge of approaching vehicle's speed and position is conjectured to be superior to only the number of queued vehicles or vehicle flow. The first vector's elements are Boolean-valued, with a one representing the presence of a vehicle and a zero representing the absence of a vehicle. The second vector's elements are real numbers and represent the vehicle's speed, normalized by the speed limit. Each element of $P$ represents a different traffic phase and all elements of $P$ are zero except for the current phase, which is one, therefore $P \in {\mathbb{B}^{|A|}}$. \par Technologies over the last decade have made gathering information required for the DTSE possible. Video cameras \cite{sivaraman2013looking} are becoming more common as sensor devices at intersections and vehicles with wireless communication capabilities (i.e., Connected Vehicles \cite{conveh}) are expected to be deployed in the near future. 
Ultimately, the DTSE is sensor agnostic, the means by which the state information is gathered, be it vision, wireless or otherwise, is irrelevant to creating the DTSE. The flexibility in generating the DTSE should be seen as an advantage of the system.

\subsection{Action Space}
After the agent has observed the state of the environment, it must choose one action from the set of all available actions. In this research, the agent's possible actions are the traffic signal phase configurations (i.e., the combination of traffic lights controlling individual lanes for the entire intersection). For simplicity and human comprehension, each action is assigned a compass direction indicating the approaching lanes' traffic signal phases (i.e., the color of the traffic signal lights) and abbreviated for brevity. For explicit clarity, a green traffic signal phase means vehicles can proceed through the intersection, yellow cautions vehicles to slow down and prepare to stop and red means vehicles should stop and not proceed through the intersection. The possible actions are North-South Green (NSG), East-West Green (EWG), North-South Advance Left Green (NSLG),  East-West Advance Left Green (EWLG). Note, for any given action, it is implied that the omitted compass direction traffic signals are red (e.g., East-West Green means that all North-South traffic signals are red).\par Formally the set of all possible actions $A$ is defined as $A = \{$NSG, EWG, NSLG, EWLG$\}$. Therefore, at time $t$, the agent chooses an action $a_t$, where $a_t \in A$. However, when an agent chooses an action, it may not be immediately enacted. To ensure safe control of the intersection, additional traffic signal phase configurations may precede the chosen action. Instead of immediately transitioning from the current traffic signal phase to the selected action, a sequence of intermediate traffic signal phases dependent on the current phase and chosen action may be necessary. All possible action transition sequences to transition from the current traffic signal to the chosen action are shown in Table \ref{action_transitions}. Note the addition of the North-South Yellow (NSY) and East-West Yellow (EWY) and All Red (R) traffic signal configurations, which cannot be chosen explicitly as actions, but are part of some traffic signal transition sequences. The yellow and red phases are necessary for safety reasons, as they slow down and stop traffic so that succeding green phases may be enacted.

\begin{table*}[]
\centering
\caption{Traffic Signal Phase Action Transitions}
\label{action_transitions}
\begin{tabular}{cllllll}
                                              & \multicolumn{1}{c}{} & \multicolumn{4}{c}{Selected Action}                                                                     &  \\ \cline{1-6}
                                              & \multicolumn{1}{c}{} & \multicolumn{1}{c}{\textbf{NSG}} & \multicolumn{1}{c}{\textbf{EWG}}& \multicolumn{1}{c}{\textbf{NSLG}} & \multicolumn{1}{c}{\textbf{EWLG}}&  \\ \cline{1-6}
\multirow{4}{*}{Current Traffic Signal Phase} & \textbf{NSG}                  & -                       & \{NSY, R\}              & \{NSY\}                  & \{NSY, R\}               &  \\
                                              & \textbf{EWG}                  & \{EWY, R\}              & -                       & \{EWY, R\}               & \{EWY\}                  &  \\
                                              & \textbf{NSLG}                 & -                       & \{NSY, R\}              & -                        & \{NSY, R\}               &  \\
                                              & \textbf{EWLG}                & \{EWY\}                 & -                       & \{EWY. R\}               & -                        &  \\ \cline{1-6}

\end{tabular}
\end{table*}

\subsection{Reward}
The final element of reinforcement learning, after the agent has observed the state of the environment $s_t$, chosen an action $a_t$, and performed it, is receiving the reward. The reward is one element that differentiates reinforcement learning from other types of machine learning; developing a state-action policy which maximizes cumulative long-term reward is what the agent seeks. Compared to other types of machine learning, in which correct actions are given by instruction, reinforcement learning has the agent evaluate actions by interacting with the environment. How to select the appropriate reward for a given task is an unanswered problem in traditional reinforcement learning\footnote{See inverse reinforcement or apprenticeship learning.}. It would be desirable if the agent could choose its own reward, instead of requiring an expert to define it, and is therefore a goal of many active researchers.\par In the context of traffic signal control, various rewards have been proposed, such as change in number of queued vehicles, change in cumulative vehicle delay and change in vehicle throughput. The reward $r_{t+1} \in \mathbb{R}$ is a consequence of enacting a selected action from a specific state. In this research, we define the reward as change in cumulative vehicle delay between actions. This allows for the reward to be positive or negative,  meaning the agent can be punished ($r_{t+1}<0$  for increase in delay) or rewarded ($r_{t+1}>0$ for decrease in delay). The use of the subscript $t+1$ is intentional, to emphasize the temporal relationship between taking action $a_t$ in state $s_t$, as the reward succeeds these two events. In addition to receiving a reward from the environment, the agent has the opportunity to observe the new state of the environment $s_{t+1}$, which was influenced by its most recent action. With this new state observation, a new action can be chosen and subsequently a new reward received. This cycle can be continued indefinitely or stopped according to some criteria, depending on the reinforcement learning task at hand.

\subsection{Agent}
In reinforcement learning, the agent is the entity that learns by interacting with the environment. We model the agent controlling the traffic signals as a deep convolutional Q-network \cite{mnih2015human}. Artificial neural networks are mathematical functions inspired by biological neural networks (i.e., brains) that are appealing for their function approximation capabilities. Many problems in machine learning can suffer from the curse of dimensionality, which is when the dimensionality of the data increases, the training and computational resources required grow exponentially. Artificial neural networks have the capability to generalize from what they have learned, weakening the problems posed by the curse of dimensionality. Convolutional neural networks are a variant of artificial neural networks inspired by biological research that emulate the architecture of the animal visual cortex \cite{hubel1962receptive}\cite{hubel1968receptive}, making them adept at perception tasks.\par Most artificial neural networks require data pre-processing, where features of the data are determined by experts. Features are measurable aspects of the data deemed important to the present machine learning task. Expert-crafted features require assumptions to be made about the data that may or may not be true. In the context of traffic signal control, examples of expert-crafted features are queue length or average vehicle flow. These features are abstractions of the individual vehicles behavior that have been extracted and deemed important by experts for solving the traffic signal control problem using reinforcement learning. However, because they are abstractions, we argue important information is lost and the potential for learning is diminished. If only queue length is used, this assumes all vehicles not in a queue are irrelevant to developing an optimal traffic signal control policy - a spurious claim. Similarly, average flow is a historical metric calculated over some time interval, yielding a very coarse approximation of the current traffic state that ignores and abstracts away useful information. Convolutional neural networks are advantageous because they develop their own features from the data. The DTSE is proposed because it is a lesser abstraction of the traffic state than queue length or average flow and the convolutional neural network can take advantage of its information rich nature.\par The depth of a deep neural network refers to the fact that there is more than one hidden computational layer of neurons. Additional layers in a network allow it to develop features of features, transforming low-level features of the data to high-level ones, potentially increasing network performance. The combination of the DTSE and the convolutional neural network allow for the creation of a truly adaptive traffic signal controller.\par The DQTSCA's architecture is first two identical networks receiving different inputs. Each network receives a different input vector from the DTSE - one the real-valued vector, the other the Boolean-valued vector. The first layer of each network is a convolutional layer with 16 filters of size 4x4 applied with stride 2 using rectifier nonlinear activation functions. The second layer of each network is a convolutional layer with 32 filters of size 2x2 applied using rectifier nonlinear activation functions. The outputs of these two networks and the vector $P$, representing the current traffic signal phase, are combined and used as input to two fully-connected layers of 128 and then 64 neurons with rectifier nonlinear activation functions. The output layer is $|A|$ (i.e., four) neurons with linear activation functions.
\par The reinforcement learning algorithm used in this research is Q-Learning \cite{watkins1992q}, which is used to develop an optimal action-selection policy. The optimal policy is achieved by using the convolutional neural network to approximate the action-value function. The action-value function maps states to action utilities (i.e., what is the value of each action from a given state). Values represent long-term reward. If an action has a high value, enacting it means reaping future reward, although potentially not immediate reward. We define the deep convolutional neural network as the action-value function $\eta : X \mapsto Y$, where $X\in S$ and $Y\in \mathbb{R}^{|A|}$, with $Y$ representing the action-values $\forall A$. At time $t$, the input $x_t$ to the network is $x_t = s_t$. The output $y_t$ is a vector containing all the action-values, with $y_{t,a_t}$ denoting the action-value of $a_t$.  After the action-value function has been sufficiently learned, the optimal policy can be determined by selecting the action with the highest value given $s_t$. The basis of Q-learning is the value iteration update (Q-update), defined in (\ref{qlearneqn}).

\begin{equation}\label{qlearneqn}
\begin{aligned}
Q(s_t, a_t) ={} & Q(s_t, a_t) +\\  
                         &  \alpha\Big( r_{t+1} + \gamma\textrm{max}_{A} Q(s_{t+1}, a_t)  - Q(s_t, a_t) \Big)\\
\end{aligned}
\end{equation}
Where the learning rate $\alpha$ controls the degree to which new action-value estimates are weighted against old estimates and the discount factor $\gamma$ determines how immediate rewards are weighted against future rewards. Both the learning rate and the discount factor are parameters of Q-learning and are $\gamma, \alpha \in [0,1]$. To use the Q-update to train the deep convolutional network, a variation of  (\ref{qlearneqn}) is used, defined in (\ref{qlearndcnn}).
\begin{equation}\label{qlearndcnn}
\begin{aligned}
x_t &= s_t\\
x_{t+1} &= s_{t+1}\\
y_t &= \eta(x_t)\\  
y_{t,a_t} &= r_{t+1} + \gamma\textrm{max}\Big(\eta(x_{t+1})\Big)
\end{aligned}
\end{equation}
After (\ref{qlearndcnn}) has been computed, the deep convolutional network can be trained and the weights $\theta$ updated using gradient descent with $x_t$ as the network input and $y_t$ as the output. We use the RMSprop \cite{tieleman2012lecture} gradient descent algorithm with an $\alpha$ of 0.00025 and a $\gamma$ of 0.95 to train the network. Once the deep convolutional neural network has sufficiently approximated the action-value function, optimal control is achieved by selecting the action with the highest value given the current state.
\par A major problem in any reinforcement learning task is the action-selection policy while learning; whether to take exploratory action and potentially learn more, or to take exploitative action and attempt to reap the most reward given what has been learned so far. The explore-exploit tradeoff is an active area of research in reinforcement learning with many proposed solutions. We implement the simple, yet effective, decreasing $\epsilon$-greedy exploration policy, which selects a random action (explore) with a probability $\epsilon$ and selects the action with the highest value (exploit) with a probability 1-$\epsilon$. The value of $\epsilon$ decreases as training epochs progress according to (\ref{egreedy}).
\begin{equation}\label{egreedy}
\epsilon_n = 1.0 - \frac{n}{N}
\end{equation}
Where $n$ is the current training epoch and $N$ is the total number of training epochs Initially, $\epsilon = 1.0$, meaning the agent exclusively explores, however, as training progresses, the agent increasingly exploits what it has learned, until it exclusively exploits.

\section{Experimental Setup and Training} \label{experiment}
All experiments were conducted using the traffic microsimulator SUMO v0.22 \cite{SUMO2012}. SUMO provides an application programming interface (API) in the Python programming language, by which custom functionality can be implemented in the traffic simulation. We used the SUMO Python API and custom code to implement the DQTSCA. The artificial neural network was implemented using Keras \cite{chollet2015keras} and Theano \cite{2016arXiv160502688short} Python libraries. Additional optimized functionality was provided by NumPy and SciPy \cite{scipy} libraries. For the DTSE parameters, we define $l$ as 75 m and $c$ as 5 m. We train for $1\thinspace600$ training epochs, where each epoch is 1.25 hours of simulated traffic. The simulations were executed on a desktop computer with a 3.40 GHz i7-2600 CPU, 8GB of RAM running Ubuntu 14.04.  The length of the agent's actions (i.e., NSG, EWG, NSLG, EWLG) are two seconds and the transition phases (i.e., R, NSY, EWY) are five seconds.\par The intersection geometry is four lanes approaching the intersection from the compass directions (i.e., North, South, East and West) connected to four outgoing lanes from the intersection. The traffic movements for each approach are as follows: the inner lane is left turn only, the two middle lanes are through lanes and the outer lane is through and right turning. All lanes are 750 meters in length, from the vehicle origin to the intersection stop line.\par The method by which vehicles are generated and released into the network greatly influences the quality of any traffic simulation. The most popular vehicle generation method is to randomly sample from a probability distribution numbers that represent vehicle headway times, or the time interval between vehicles. This research does not break from this method entirely, however we strive to implement a nuanced version which better models real-world traffic. Empirical research has shown that different vehicle flow rates are suitably approximated by different probability distributions \cite{riccardo2012empirical}\cite{maurya2015speed}. Instead of using a negative exponential distribution for all flow rates and modifying its rate parameter, we use different distributions for different flow rates, shown in Table \ref{veh_gen}. The Inverse Weibull distribution is used for generating left and right turning traffic and the Burr distribution is used for generating through traffic.

\begin{table*}[]
\centering
\caption{Vehicle Generation Distributions by Flow Rate}
\label{veh_gen}
\begin{tabular}{ccc}
\hline
\textbf{Flow Rate (Vehicles/Hour)} & \textbf{Distribution}    & \textbf{Parameters $(\alpha, \beta)$} \\ \hline
0-150     & Inverse Weibull \cite{riccardo2012empirical}& (0.65, 5.8) \\
250-450   & Burr \cite{maurya2015speed}          & (1.4, 5.9)  \\ \hline
\end{tabular}
\end{table*}
\par The agent is trained using a biologically inspired process known as experience replay \cite{mcclelland1995there}\cite{o2010play}\cite{lin1993reinforcement}. Instead of training after every individual state, action, reward, state sequence, the agent stores the experience, defined $e_t = (s_t, a_t, r_{t+1}, s_{t+1} )$, in an experience memory $M$ for periodic, randomized batch training. The training pseudocode is presented in Algorithm 1 and 2. This research takes advantage of the multithreading capabilities of modern computers, running multiple traffic simulations in parallel. Each thread is running Algorithm 2 and generating different experiences for use in the experience replay.\\

\begin{minipage}{\linewidth}
\vspace{4mm}
\textbf{Algorithm 1: Deep reinforcement learning traffic signal control agent experience replay}\\
Initialize neural network agent $\eta$ with random weights $\theta$ on main agent\\
Copy main agent weights $\theta$ to all thread agents\\
\textbf{For} \textit{epoch}=1 to $N$ \textbf{do}\\
\phantom{x}Copy main agent weights $\theta$ to all thread agents\\
\phantom{x}In parallel run Algorithm 2 on threads\\
\phantom{xx}\textbf{While} all threads not finished \textbf{do}\\
\phantom{xxx}\textbf{If} \textit{buffer} == \textit{batch\_size} \textbf{do}\\
\phantom{xxxx}Append \textit{buffer} to $M$, clear \textit{buffer}\\
\phantom{xxxx}Randomly sample \textit{batch\_size} experiences,\\
\phantom{xxxx}from $M$\\
\phantom{xxxx}Batch train main agent using (\ref{qlearndcnn})\\
\phantom{x}\textbf{If} \textit{epoch}\thinspace mod(\textit{exp\_refill}) == 0 \textbf{do}\\
\phantom{xx}Clear $M$\\
\phantom{xx}\textbf{While} len($M$) $<$ \textit{min\_size} \textbf{do}\\
\phantom{xxx}In parallel run Algorithm 2 on threads\\
\vspace{4mm}
\phantom{xxx}Append \textit{buffer} to $M$\\
\textbf{Algorithm 2: Thread Traffic Simulation }\\
\phantom{}\textbf{For} $t$=1 to \textit{sim\_len} \textbf{do}\\
\phantom{x}Observe DTSE, $s_{t}$\\
\phantom{x}Select random action $a_t$ with probability $\epsilon$,\\
\phantom{x}else select $a_t = \textrm{max} \eta(s_{t})$\\
\phantom{x}Implement selected $a_t$, increment simulation,\\
\phantom{x}observe reward $r_{t+1}$ and $s_{t+1}$,\\
\phantom{x}\textbf{If} len($M$) == \textit{max\_size} \textbf{do}\\
\phantom{xx}delete $M$[0]\\
\phantom{x}Append $e_t = (s_{t}, a_{t}, r_{t+1}, s_{t+1})$ to \textit{buffer}\\
\end{minipage}
The \textit{buffer} in Algorithm 1 and 2 temporarily stores the most recent experiences until it reaches \textit{batch\_size}, at which point it is appended to \textit{M} and cleared. The \textit{max\_size} and \textit{min\_size} are respective upper and lower limits of \textit{M}. Training can only begin after \textit{M} has at least \textit{min\_size} experiences. The oldest experience is deleted when \textit{M} has \textit{max\_size} elements. In our research, we use a \textit{batch\_size} of 16, a \textit{max\_size} of 500\thinspace000 and a \textit{min\_size} of 50\thinspace000. We found learning improved when we periodically cleared \textit{M} and refilled it with new experiences every \textit{exp\_refill} epochs, where \textit{exp\_refill} is $200$. The \textit{sim\_len} is 4\thinspace500 timesteps.\par We developed a shallow neural network TSCA to compare against our proposed DQTSCA. The shallow traffic signal control agent (STSCA) has one hidden layer with 64 neurons using the sigmoid activation function and four neurons with linear activation functions for its output layer. The state space of the STSCA is two vectors, the first containing elements that represent the number of queued vehicles at each intersection approach (i.e., North, South, East and West) and the second the current traffic signal phase vector \textit{P}. The action space and reward are the same as the DQTSCA. The STSCA is trained using the same number of epochs, action selection policy and gradient descent algorithm as the DQTSCA. However, the traditional agent does not use experience replay, it trains using (\ref{qlearndcnn}) after every state, action, reward, state sequence.
\section{Results and Discussion} \label{results}
\begin{figure}
\centering
\includegraphics[scale=0.43 ]{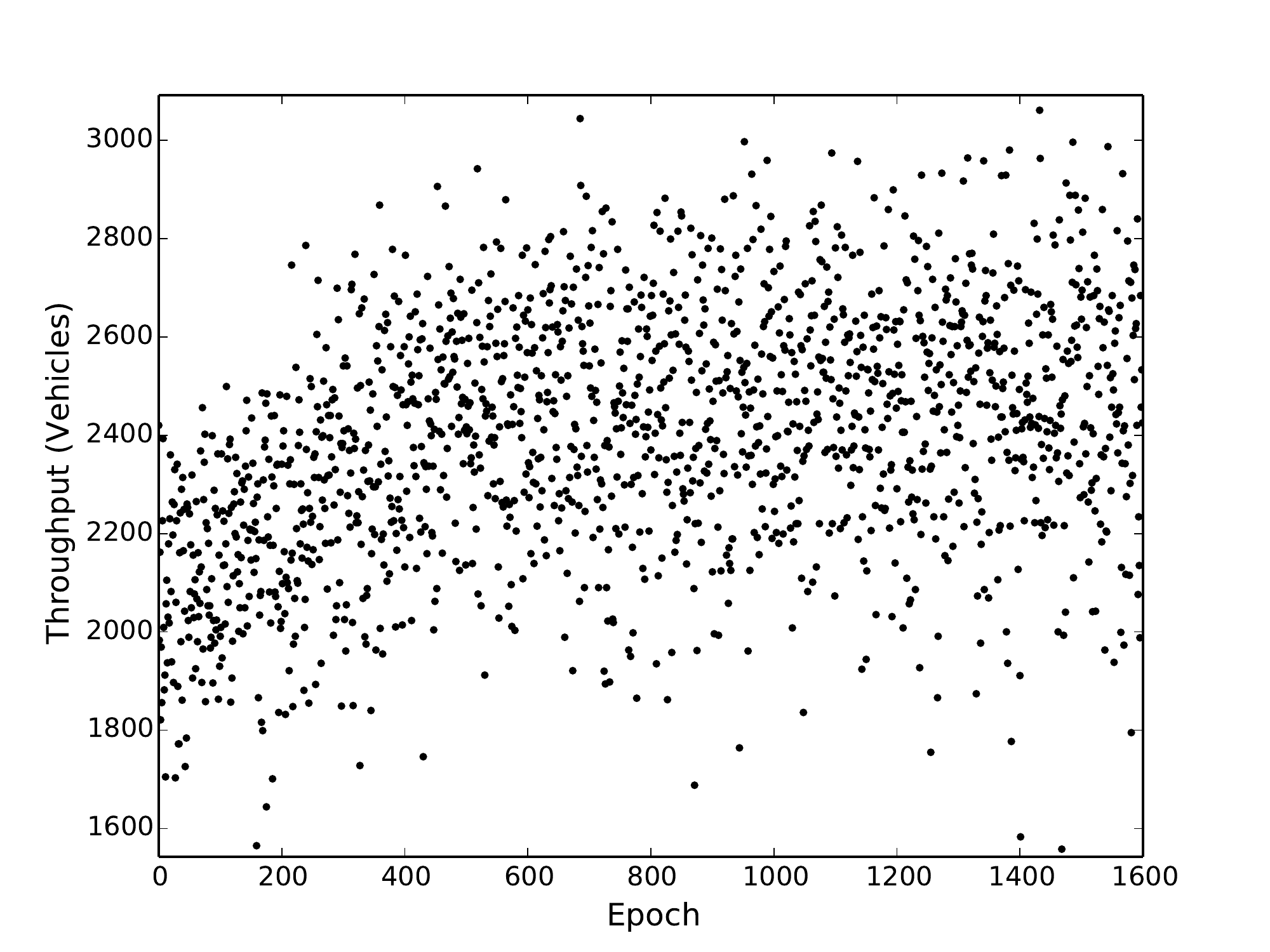}
\caption{Intersection throughput while training.}
\label{fig:through}
\end{figure}

\begin{figure}
\centering
\includegraphics[scale=0.43 ]{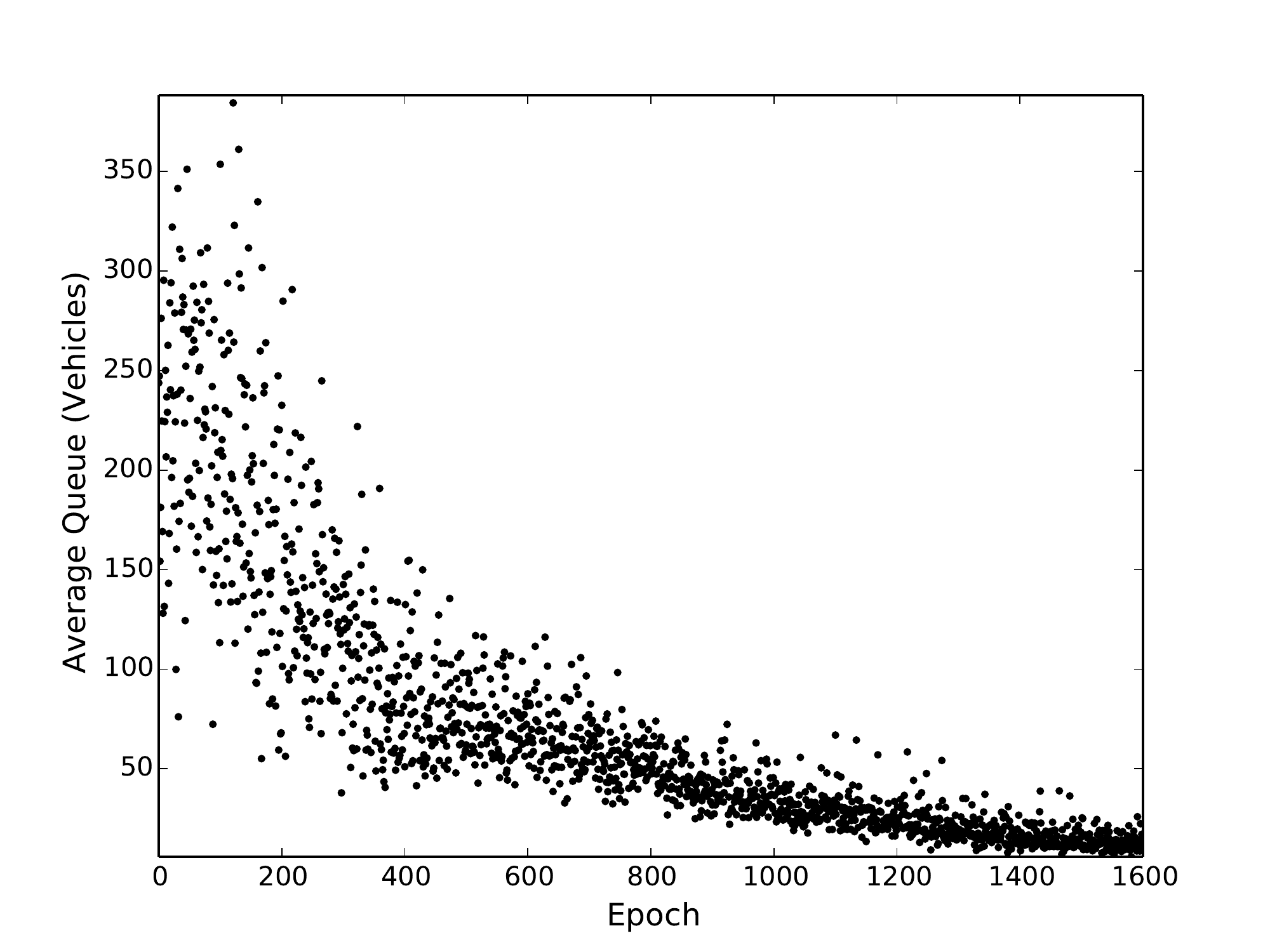}
\caption{Average intersection queue while training.}
\label{fig:queue}
\end{figure}

\begin{figure}
\centering
\includegraphics[scale=0.43 ]{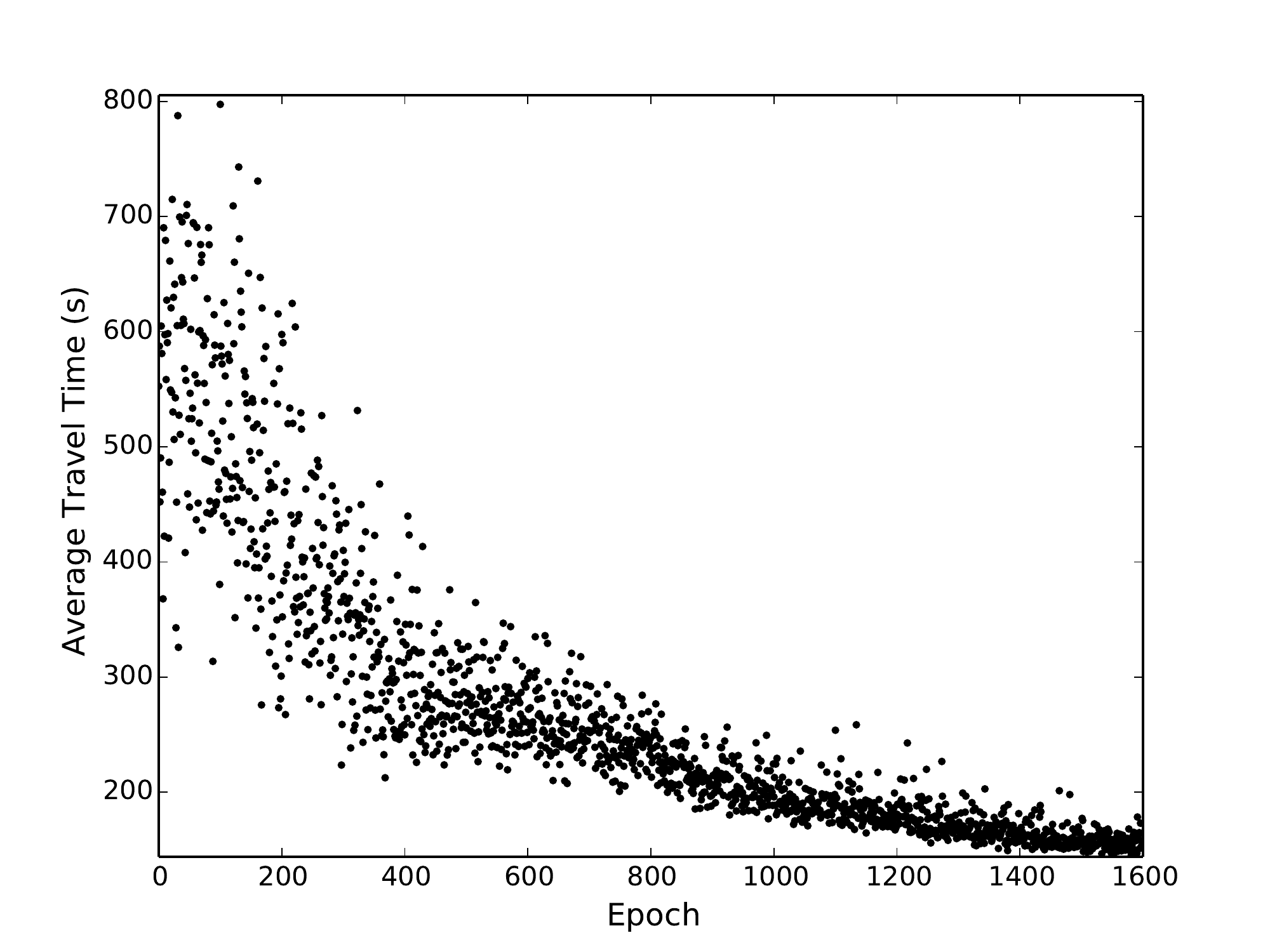}
\caption{Average travel time of vehicles while training.}
\label{fig:traveltime}
\end{figure}

\begin{figure}
\centering
\includegraphics[scale=0.43 ]{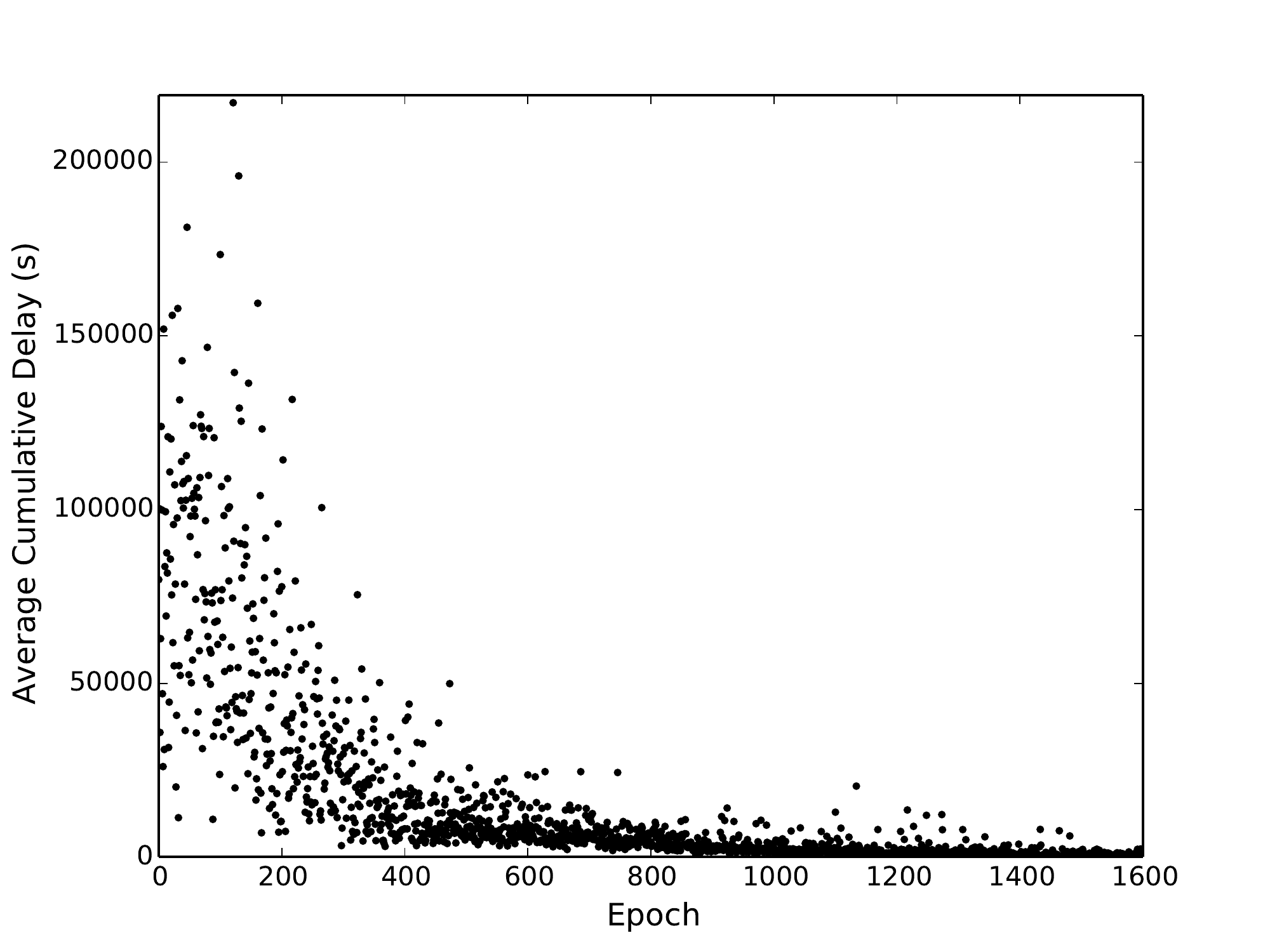}
\caption{Average cumulative delay of vehicles while training.}
\label{fig:delay}
\end{figure}

\begin{figure}
\centering
\includegraphics[scale=0.43 ]{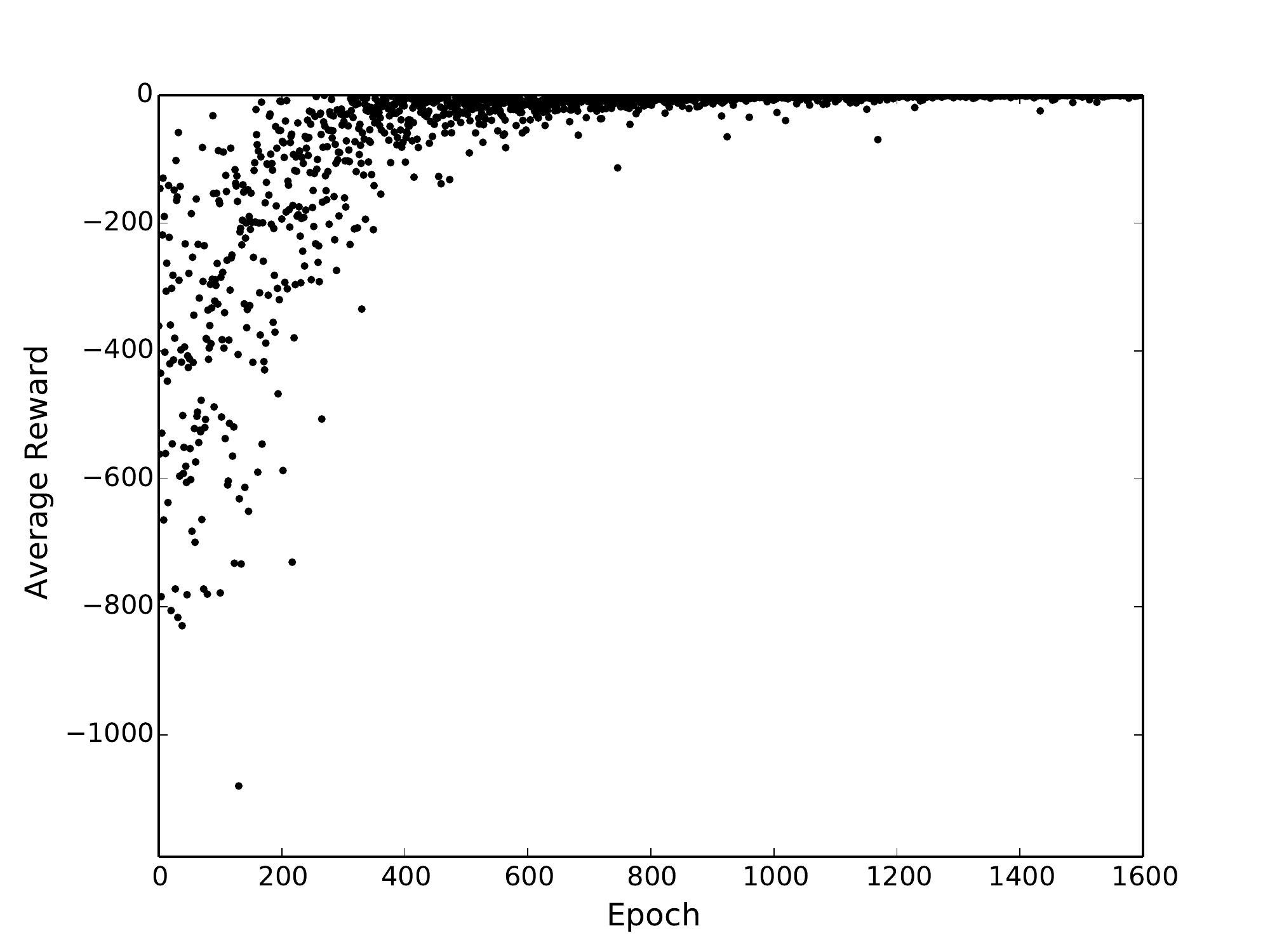}
\caption{Average reward of DQTSCA while training.}
\label{fig:reward}
\end{figure}

\begin{figure}
\centering
\includegraphics[scale=0.43 ]{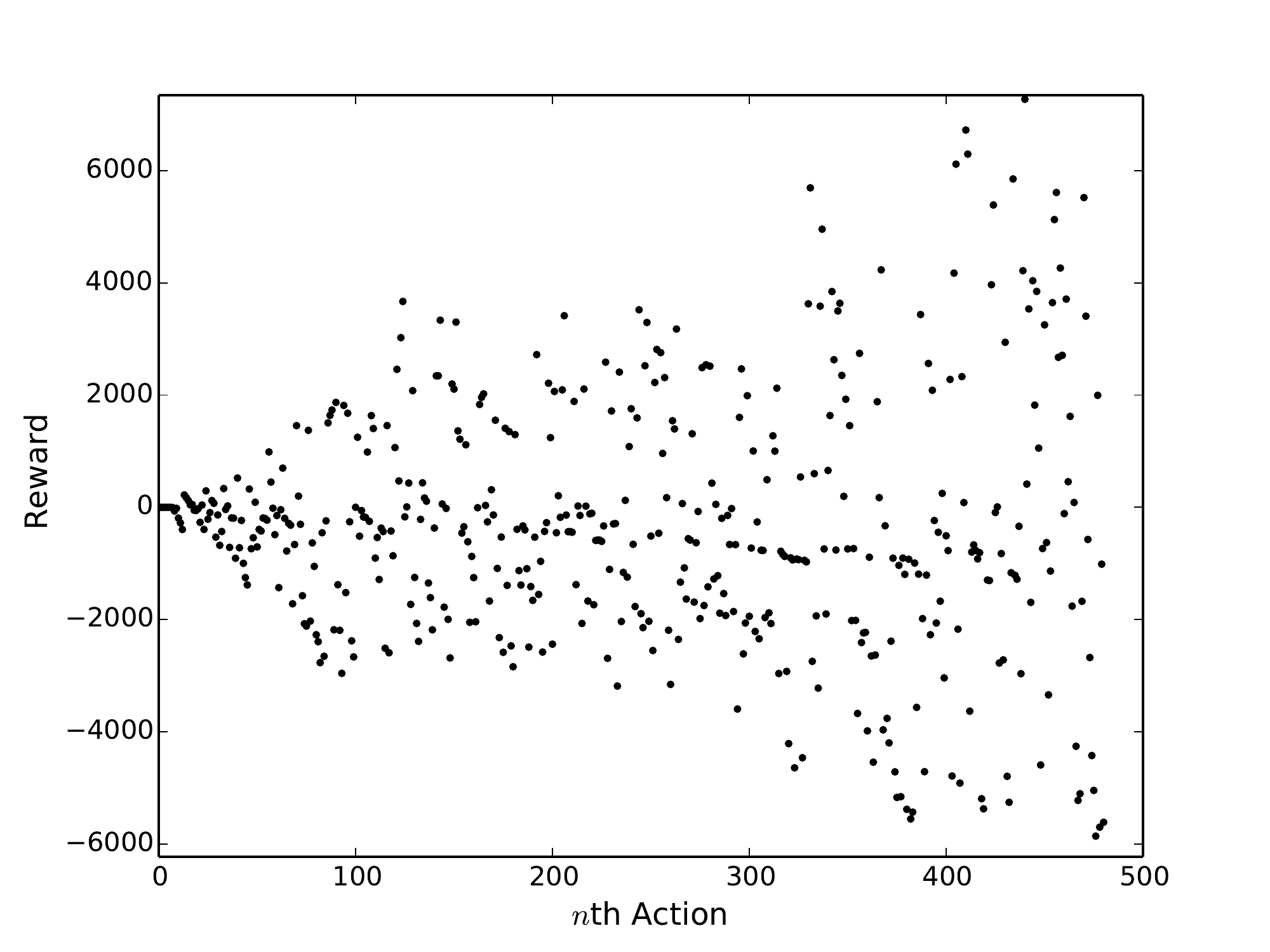}
\caption{Reward of DQTSCA in an epoch while taking only exploratory action early in training.}
\label{fig:explore_reward}
\end{figure}

\begin{figure}
\centering
\includegraphics[scale=0.43 ]{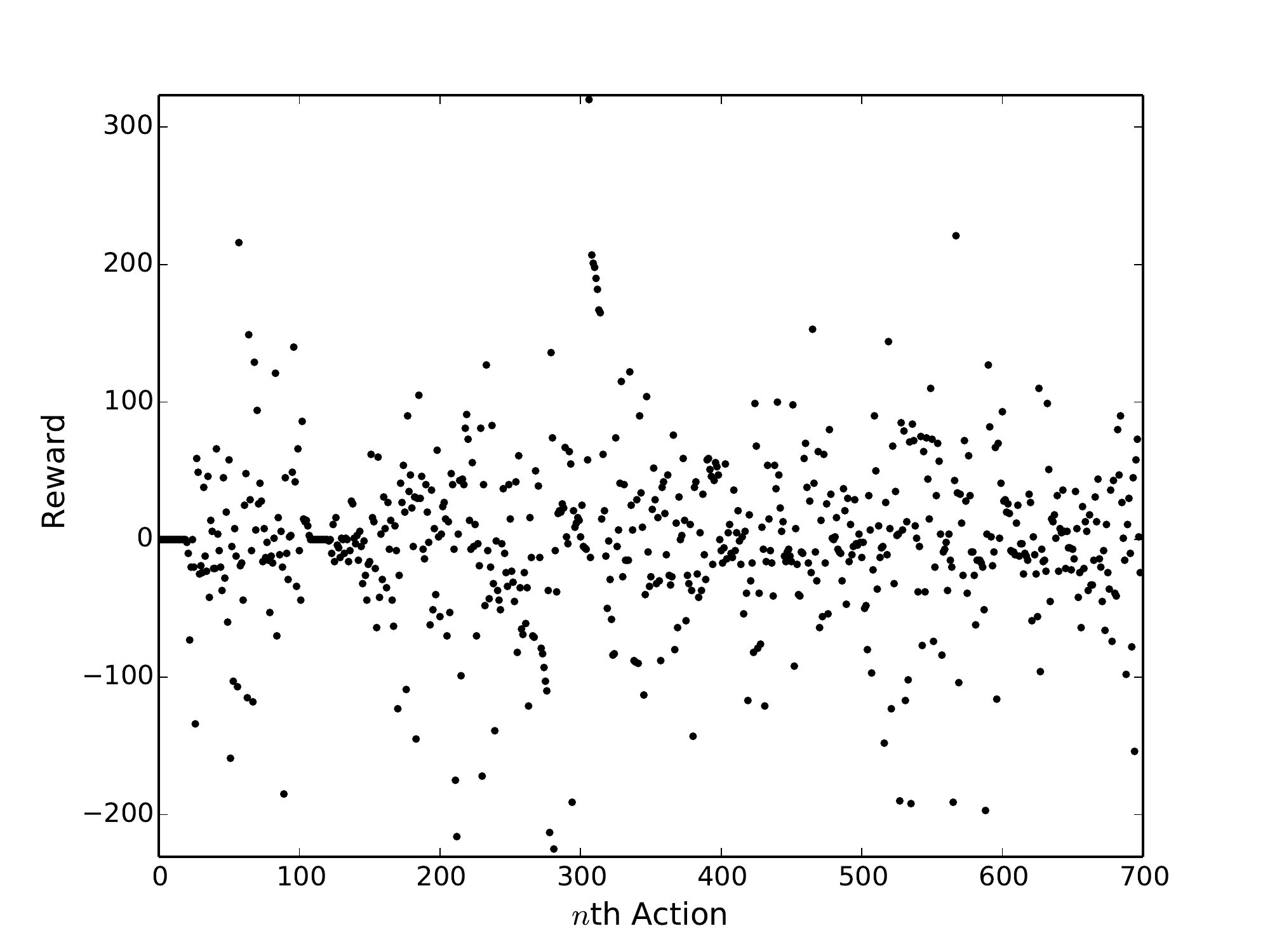}
\caption{Reward of DQTSCA in an epoch while taking only exploitative action after training completed.}
\label{fig:exploit_reward}
\end{figure}

The performance of the proposed DQTSCA was assessed with respect to common traffic metrics: throughput, queue length, travel time and cumulative delay. The performance of the agent with respect to the traffic metrics while learning can be seen in Figures \ref{fig:through}, \ref{fig:queue}, \ref{fig:traveltime}, and \ref{fig:delay}. The agent's performance with respect to achieving reward while learning can be seen in Fig. \ref{fig:reward}. The agent's action-reward performance during one epoch is also shown, in Fig. \ref{fig:explore_reward} exclusively exploring initially in training and exclusively exploiting after training in Fig. \ref{fig:exploit_reward}. Initially, while learning, the agent is predominantly exploring (i.e., taking random actions), attempting to learn the action-value function. While exploring, the agent's performance with respect to the traffic metrics exhibits high variance and it achieves negative reward (i.e., punishment). Because of the agent's actions, many vehicles are queued, unnecessarily delayed and the overall throughput is low. As the epochs progress, the agent has better learned the action-value function and can begin selecting exploitative actions instead of exploratory ones. The decreasing exploration rate is reflected in improved performance with respect to all four metrics and higher reward - evidence the agent has learned. Not only does the DQTSCA perform better as training progresses, convergent behavior emerges, as the variance in its performance decreases.\par The agent's behavioral change before and after training can be seen in Figures \ref{fig:explore_reward} and \ref{fig:exploit_reward}. These figures show rewards as a consequence of each action taken within one epoch (i.e., 1.25 hours of simulated traffic). In Fig. \ref{fig:explore_reward}, the agent is taking random actions with no consideration for reward, reflected as unstable and divergent rewards. The key observation is that the rewards, positive or negative, increase in magnitude as the epoch progresses because the agent is taking random, exploratory actions. Comparing Fig. \ref{fig:exploit_reward} with Fig. \ref{fig:explore_reward}, it is apparent the agent is acting differently. In Fig. \ref{fig:exploit_reward}, the rewards are an order of magnitude smaller and stable as the epoch progresses, with no divergence near the end of the epoch as in Fig. \ref{fig:explore_reward}, because the agent is enacting an exploitative policy. These observations are supported quantitatively, computing the average and standard deviation $(\mu,\sigma)$ of the reward for each epoch, Fig. \ref{fig:explore_reward} has $(-347, 2\thinspace220)$ and Fig. \ref{fig:exploit_reward} has $(-0.485, 59.6)$.\par A comparison of the proposed DQTSCA with the STSCA can be seen in Table \ref{agent_compare}. The data in Table \ref{agent_compare} is computed from the last 100 training epochs of each agent, where the agents are taking exploitative action $>$$93\%$ of the time. Although four traffic metrics are considered, cumulative delay is the only metric the agent can tangibly interact with, as change in cumulative delay is its reward function. The DQTSCA achieves an $82\%$ reduction in the average cumulative delay compared to the STSCA. The difference in this key metric provides evidence that the DQTSCA has learned a control policy superior to the STSCA. Comparing the other traffic metrics, there is no difference in the throughput, but the DQTSCA reduces the average queue length by $66\%$ and the average travel time by $20\%$ compared to the STSCA. The DQTSCA outperforms the STSCA in three of the four metrics, due to the use of the DTSE and its deep architecture. Future work should investigate a throughput reward function and compare the two agents performance, as it is the only metric where the agents perform equally.\par A limitation of this research is we did not consider how fair the agent's policy is. A fair traffic signal controller would ensure all vehicles are given equal priority to traverse the intersection, however this may be in conflict with optimizing certain traffic metrics, such as minimization of delay or maximization of throughput. A balance between fairness and optimality could be achieved with the appropriate reward function, which should be the subject of future research.
\begin{table*}[]
\centering
\caption{STSCA and DQTSCA Traffic Metrics}
\label{agent_compare}
\begin{tabular}{lll}
\hline
\textbf{Traffic Metric} ($\mu$, $\sigma$, $n=100$) & \textbf{STSCA}      & \textbf{DQTSCA}   \\ \hline
Throughput (Vehicles) & (2\thinspace452, 257)  & (2\thinspace456, 248)  \\
Queue (Vehicles)                 & (33, 23)     & (13, 9)     \\
Travel Time (s)                  & (197, 107)  & (157, 49)  \\
Cumulative Delay (s)             & (4\thinspace085, 5\thinspace289) & (719, 1\thinspace048) \\ \hline
\end{tabular}
\end{table*}
\section{Conclusion} \label{conclusion}
We proposed, developed and tested a DQTSCA in a traffic microsimulator. The results show deep learning can be applied to traffic signal control with improved performance compared to traditional methods.\par Future work in this area can extend the agent's control to all traffic signals, including the yellow and red phases. Currently, the agent has no capability to control the yellow or red phases, they only exist in the transitional sequences between agent actions. However, it is obvious that situations exist where a dynamic yellow or red phase is desirable. For instance, a vehicle does not decelerate to a yellow phase and accelerates through the intersection; extending the yellow phase until all vehicles have either cleared the intersection or are otherwise decelerating is prudent. We hypothesize the means to accomplish this with a TSCA trained through reinforcement learning would be to change the reward function so that it yielded high reward when vehicles decelerate and are not traversing the intersection.\par The use of the DTSE may also allow for training a TSCA to control intersections of various lane configurations without retraining. For example, first train the agent using the DTSE on a four lane intersection. Then it could be used to control a two lane intersection by setting all of the elements of the two 'missing' lanes to zero. The DTSE may allow for a widely applicable TSCA without retraining for different intersection geometries.\par Additional research should also increase the complexity of the traffic network and apply the DTSE and deep architecture to multiple TSCA. These ideas will be explored in future endeavors by the authors.

\section*{Acknowledgment}

The authors would like to acknowledge the authors of the various programming libraries used in this research. The authors would also like to acknowledge the comments and support of friends, family and colleagues.

\ifCLASSOPTIONcaptionsoff
  \newpage
\fi



\bibliographystyle{IEEEtran}
\bibliography{bare_jrnl}


%



%

\begin{IEEEbiography}[{\includegraphics[width=1in,height=1.25in,clip,keepaspectratio]{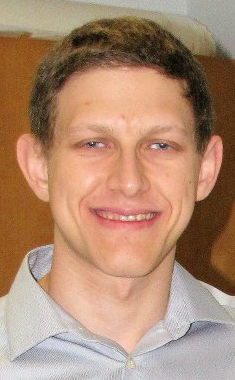}}]{Wade Genders}
earned a Software B.Eng. \& Society in 2013 and Civil M.A.Sc. in 2015 from McMaster University. He is currently pursuing a Ph.D. at McMaster University in the Department of Civil Engineering. His research topics include traffic signal control, intelligent transportation systems, machine learning and artificial intelligence.
\end{IEEEbiography}

\begin{IEEEbiography}[{\includegraphics[width=1in,height=1.25in,clip,keepaspectratio]{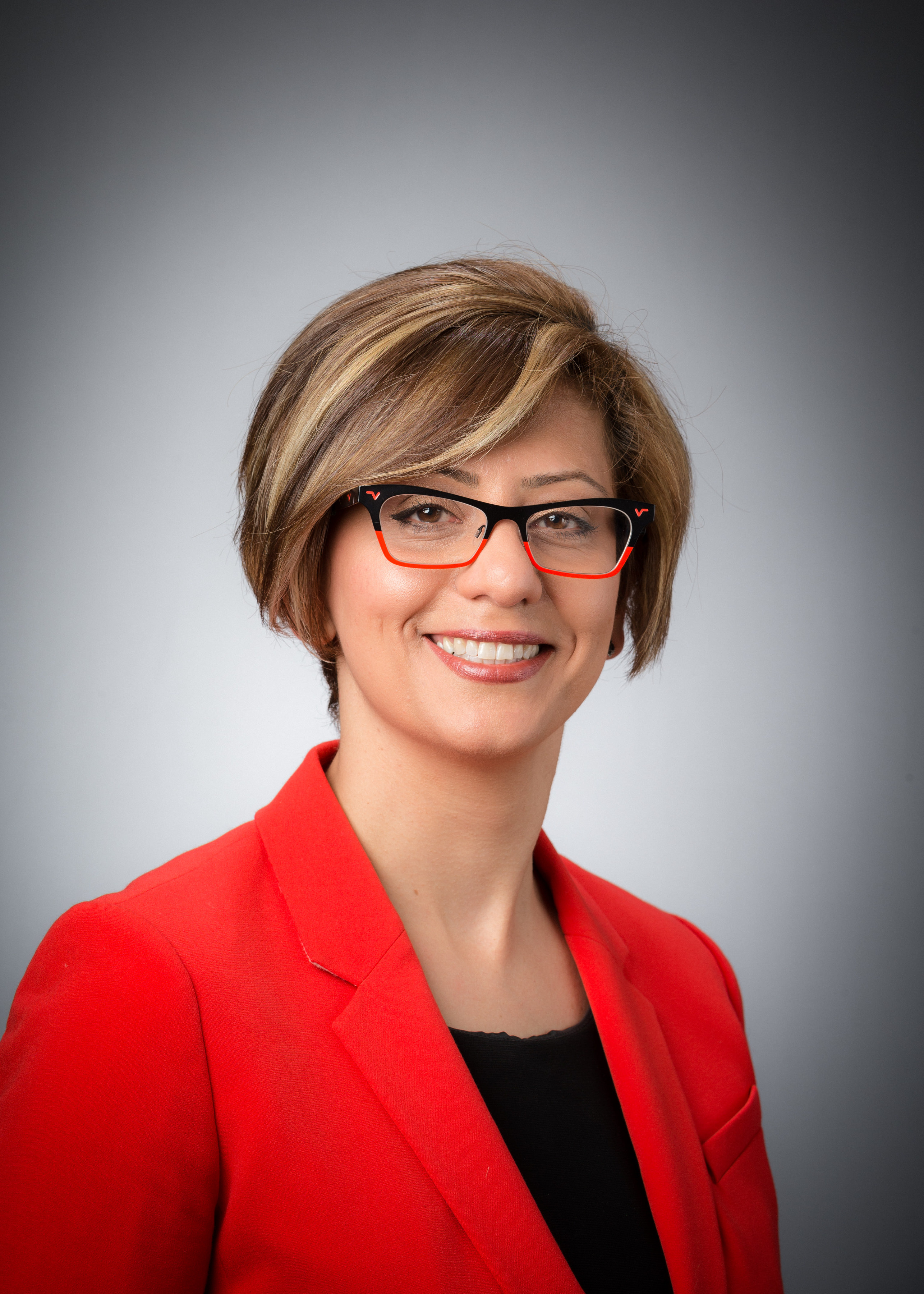}}]{Saiedeh Razavi}
is the inaugural Chair in Heavy Construction, and Associate Professor at the Department of Civil Engineering at McMaster University.  Dr. Razavi has a multidisciplinary background and considerable experience in collaborating and leading national and international multidisciplinary team-based projects in sensing and data acquisition, sensor technologies, data analytics, data fusion and their applications in safety, productivity, and mobility of transportation, construction, and other systems. She combines several years of industrial experience with academic teaching and research. Her formal education includes degrees in Computer Engineering (B.Sc), Artificial Intelligence (M.Sc) and Civil Engineering (Ph.D.). Her research, funded by Canadian council (NSERC), as well as the ministry of Transportation of Ontario, focuses on connected and automated vehicles, on smart and connected work zones and on computational models for improving safety and productivity of highway construction.  Dr. Razavi brings together the private and public sectors with academia for the development of high quality research in smarter mobility, construction and logistics. She has received several awards including McMaster's Student Union Merit Award for Teaching, the Faculty of Engineering Team Excellent Award, and the Construction Industry Institute best poster award. 
\end{IEEEbiography}





\end{document}